\documentclass[sn-mathphys]{sn-jnl}  

\usepackage{color}
\usepackage{cancel}

\usepackage{algorithm}
\usepackage{algpseudocode}
\usepackage{bm}

\usepackage{multirow, pifont}
\usepackage{amssymb}
\usepackage{amsfonts}
\usepackage{bbm}
\usepackage{hyperref}

\jyear{2021}%

\theoremstyle{thmstyleone}%
\theoremstyle{thmstyletwo}%

\theoremstyle{thmstylethree}%

\newcommand{\cmark}{\ding{51}}%
\begin{document}

\title[Dilation-Erosion for Single-Frame Supervised Temporal Action Localization]{Dilation-Erosion for Single-Frame Supervised Temporal Action Localization}




\author[1]{\fnm{Bin} \sur{Wang}}\email{lingjun@njust.edu.cn}

\author*[1]{\fnm{Yan} \sur{Song}}\email{songyan@njust.edu.cn}
\author[2]{\fnm{Fanming} \sur{Wang}}\email{527288991@qq.com}
\author[1]{\fnm{Yang} \sur{Zhao}}\email{\{zhaoyang, shuxb, ruiyan\}@njust.edu.cn}
\author[1]{\fnm{Xiangbo} \sur{Shu}}
\author[1]{\fnm{Yan} \sur{Rui}}

\affil*[1]{ \orgname{Nanjing University of Science and Technology},
\orgaddress{ \postcode{210094}, \state{Jiangsu}, \country{China}}}

\affil[2]{\orgname{Nanjing Xidao Culture Communication
Ltd}, \orgaddress{ \postcode{210094}, \state{Jiangsu}, \country{China}}}




\abstract{
To balance the annotation labor and the granularity of supervision, single-frame annotation has been introduced in temporal action localization. 
It provides a rough temporal location for an action but implicitly overstates the supervision from the annotated-frame during training, leading to the confusion between actions and backgrounds, i.e., action incompleteness and background false positives.
To tackle the two challenges, in this work, we present the Snippet Classification model and the Dilation-Erosion module.
In the Dilation-Erosion module, we expand the potential action segments with a loose criterion to alleviate the problem of action incompleteness and then remove the background from the potential action segments to alleviate the problem of action incompleteness.
Relying on the single-frame annotation and the output of the snippet classification, the Dilation-Erosion module mines pseudo snippet-level ground-truth, hard backgrounds and evident backgrounds, which in turn further trains the Snippet Classification model. 
It forms a cyclic dependency.
Furthermore, we propose a new embedding loss to aggregate the features of action instances with the same label and separate the features of actions from backgrounds.
Experiments on THUMOS14 and ActivityNet 1.2 validate the effectiveness of the proposed method. 
Code has been made publicly available\footnote{\url{https://github.com/LingJun123/single-frame-TAL}}.
}


\maketitle

\begin{figure}[ht]
\centering
\includegraphics[scale=0.75]{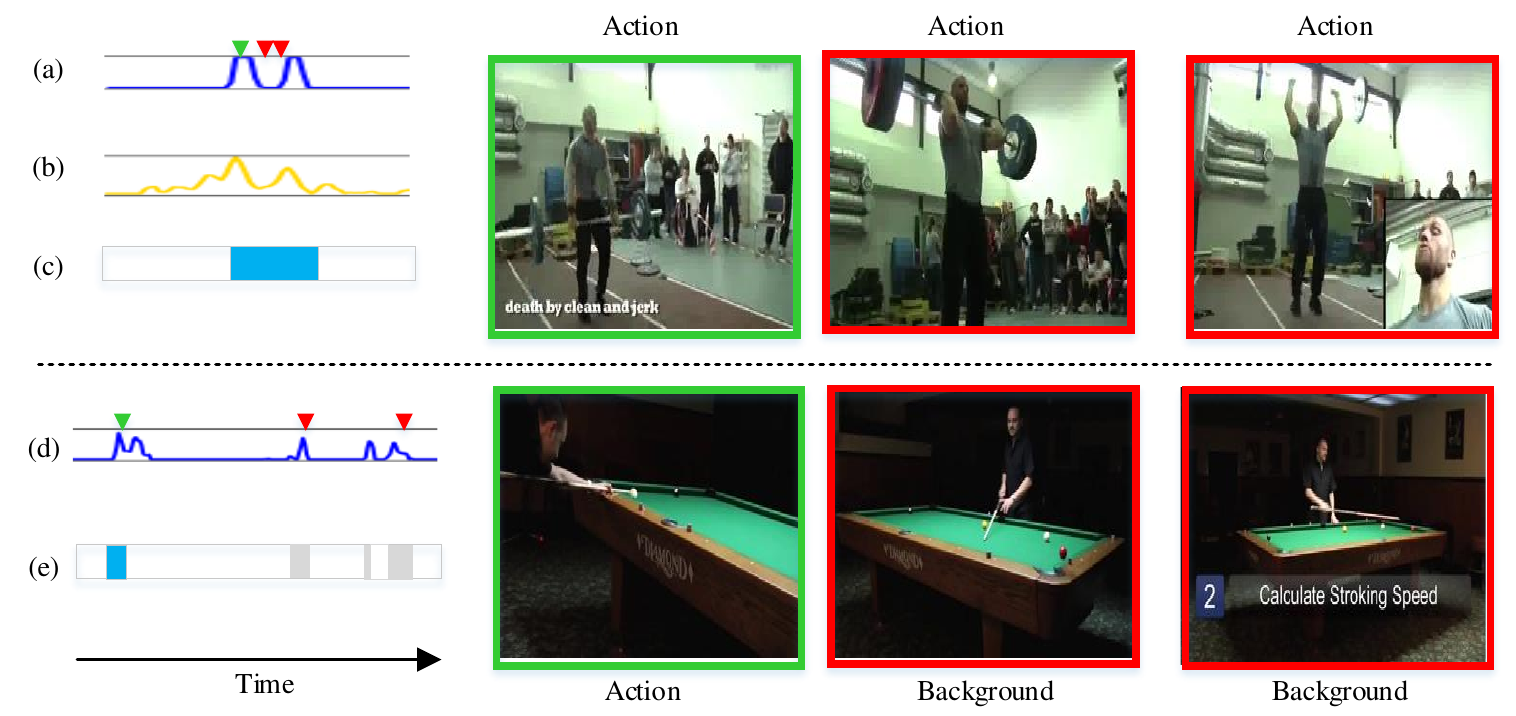}
\caption{Two visualization examples illustrate the two challenges: action incompleteness and background false positives. The green bounding boxes represent the labeled frames and the red bounding boxes represent the misjudged frames. \textbf{Top:} an example of action incompleteness. 
(b) The actionness score. 
(c) Ground-truth of an action instance of \emph{Clean and Jerk}. 
\textbf{Bottom:} an example of background false positives. 
(d) The CAM. 
(e) The blue part represents the ground truth, and the gray parts represent background false positives. 
 Best viewed in color.}
\label{fig:intro}
\end{figure}

\section{Introduction}\label{intro}

Temporal action localization (TAL), which aims at determining action categories and temporal boundaries (i.e., start and end time) of action instances occurring in videos, has received widespread attention from the community. Recently, fully-supervised temporal action localization has made significant progress~\cite{10.1007/978-3-030-01225-0_1, 9298475, 9171561, 8954471}, using frame-level annotation: action category, start and end time. However, for large-scale untrimmed videos, annotating in this way is not only cumbersome but also subjective. To relieve the pressure of annotating, weakly-supervised temporal action localization (WTAL) has been extensively studied~\cite {8953341, paul2018w, shou2018autoloc, yu2019temporal, lee2020background}, using only video-level annotation. Due to the absence of frame-level annotations, most WTAL methods considered a video as a bag of frames and fed it into video-level classification networks, leading to:  i) action incompleteness~\cite{yu2019temporal, 8237643, su2018cascaded, 10.1145/3240508.3240511}, which means that only the discriminative parts in actions can be detected;  ii) background false positives~\cite{lee2020background, 10.1007/978-3-030-58526-6_43, 9009568}, which are the backgrounds classified as actions by the model, i.e., hard background. To tackle the two challenges, adding additional supervision has become a trend, e.g., movie scripts~\cite{4587756,6751394}, the count annotation of action category~\cite{9008791}, single-frame annotation of action instances~\cite{10.1007/978-3-030-58548-8_25}, and segment-level annotation of action instances~\cite{2020Weakly}. It is worth mentioning that single-frame annotation is able to provide temporal information compared with count annotation and it is much easier to achieve than segment-level annotation. 
Single-frame annotation indicates the coarse temporal location of the action instance, but it tends to force the network to detect the frames similar to the annotated frame. 
The advantages and challenges of single-frame annotation are analyzed as follows:

\subsection{Advantages of single-frame annotation}
Firstly, action instances could be roughly located easily with the help of single-frame annotation. Due to the fact that the frames in an action instance are similar and continuous~\cite {9093620, paul2018w}, coarse temporal boundaries could be determined by mining the segments which contain the annotated frame and whose frames are similar to it. Secondly, due to the absence of the temporal annotation, WTAL methods determine action segments by relying on the prediction scores, resulting in an irreconcilable problem: background false positives. Comparatively, single-frame annotation is able to play a crucial part in recognizing background false positives. If a detected segment does not contain the annotated frame and is not close to it temporally, it is likely to be a hard background. Examples of hard background are shown in Fig. \ref{fig:intro}.

\subsection{Challenges of single-frame annotation}
Although single-frame annotation provides a rough temporal position of an action instance, it implicitly places too much emphasis on the annotated frame during localization. It might aggravate the confusion between actions and backgrounds. As a matter of fact, there may be a variety in the frames of an action instance. Besides, we could not deny the fact that there are situations where the backgrounds and the annotated frames are very similar. In another word, the above mentioned issues of WTAL might still exist. However, unlike weak-supervision based methods that regard TAL as a MIL task, single-frame supervised TAL faces the two challenges because of the over-learning of annotated frames. The first challenge is how to completely localize action instances without full annotations. For example as shown in Fig \ref{fig:intro}, in \emph{Clean and Jerk}, the beginning of an action instance is labeled so the model outputs high scores for frames (beginning and ending) similar to the labeled frame and low scores for the dissimilar frames (middle). The second challenge is how to suppress false positives. Because of the complexity of the background, they might be similar to the annotated frames in many scenes (e.g., \emph{Billiards}) in Fig \ref{fig:intro}. In the circumstances, backgrounds also get high actionness scores and bring in background false positives. Unfortunately, SF-Net~\cite{10.1007/978-3-030-58548-8_25} lacks extensive exploration for these two situations.

In this work, we propose a novel framework, as shown in Fig. \ref{fig:framework}, which consists of two core parts: Snippet Classification model and Dilation-Erosion module. Specifically, the Snippet Classification model is able to generate Class Activate Map (CAM), actionness score and class-agnostic attention, which are used to infer action segments in a video and the following processing of the Dilation-Erosion module.
To make full use of single-frame annotations and avoid the two challenges, we introduce the Dilation-Erosion module including Segment Dilation and Segment Erosion. In the Segment Dilation step, we use the Class Activate Map (CAM) to mine potential action segments and use the actionness score to expand the potential action segments which contain the annotated frames. This step discovers sufficiently large potential action segments without missing the action parts. In the Segment Erosion step, we propose  the Seed Temporal Growing (STG) algorithm to refine action parts and further remove backgrounds, including hard backgrounds and evident backgrounds. After this step, we could obtain pseudo snippet-level labels, hard backgrounds and evident backgrounds, which in turn supervise the training of the Snippet Classification model. It forms a cyclic dependency. 
Our method is able to mine the evident background as well as the hard background, while SF-Net~\cite{10.1007/978-3-030-58548-8_25} just focuses on the former.
Furthermore, we propose a new embedding loss, which is able to aggregate the features of action instances with the same label and separate the features of actions from backgrounds.
To minimize the impact of initialization, we first train the framework on the single-frame supervision and then continue training it on the pseudo snippet-level labels. During the two-stage learning process, the less discriminative action segments can be gradually detected, while the background false positives can be gradually removed.

Our contributions are three-fold:

\begin{itemize}

\item [1)] 
We propose a new framework, which consists of two core parts: Snippet Classification model and Dilation-Erosion module. 
The two parts form a cyclic dependency. As the iteration proceeds, the less discriminative action segments can be gradually detected, while the background false positives can be gradually removed.

\item [2)]
We introduce a new embedding loss, which aggregates the features of action instances with the same label and separates the features of actions from backgrounds.

\item [3)]
Experiments show that suppressing background false positives by mining hard backgrounds can significantly improve the performance. 
In particular, our method outperforms the SF-Net~\cite{10.1007/978-3-030-58548-8_25} by 5.8\% (mAP @ IoU=0.5), which further proves the effectiveness of the proposed method.

\end{itemize}

The remaining of the paper is organized as follows. Section~\ref{related} reviews literatures related to single-frame temporal action localization, including fully-supervised and weakly-supervised temporal action localization, and point supervision as well. We introduce the proposed method and the details of the training and inference phases in Section~\ref{method}. Section~\ref{exp} gives the experimental setting, the comparison with SOTA methods and the ablation experiments. Finally, the conclusion is drawn in Section~\ref{conclusion}.

\section{Related Work}
\label{related}

\subsection{Fully-supervised temporal action localization}
The task of fully-supervised TAL is to detect the boundaries and categories of action instances based on frame-level annotations. Early works~\cite{6638088, 6909495, 7780706} utilized hand-crafted features~\cite{wang2013dense,laptev2005space} with traditional learning methods~\cite{ye2019nonpeaked,cortes1995support,fu2020learning} to identify the temporal boundaries of action instances in videos. For example, Yuan \emph{et al.}~\cite{7780706} proposed a Pyramid of Score Distribution Feature (PSDF) to capture the motion information at multiple resolutions centered at each detection window and used the LSTM~\cite {Kalchbrenner2016GridLS} network to perform temporal action localization on the PSDF features.

Inspired by the great success of deep learning in the field of object localization~\cite {girshick2014rich, 8879668, 9305976} and action recognition~\cite{yan2020social,yan2018participation,yan2020higcin}, fully-supervised temporal action localization based on deep learning has been extensively studied.
Early works generated action proposals with the help of anchors, and then utilized boundary regressors and action classifiers to refine and classify the proposals respectively. The two-stage methods~\cite{shou2016temporal, 8237654, shou2017cdc, gao2017cascaded, 8099821, 8578222} adopted proposal-plus-classification, that is, the proposals are generated and then classified. Compared with two-stage methods, one-stage methods used anchors to directly predict action instances. Lin \emph{et al.}~\cite{lin2017single} proposed a framework called SSAD which predicted multiple-scale action instances without the proposal generation step. Long \emph{et al.}~\cite{long2019gaussian} developed a novel framework called GTAN which introduced the Gaussian kernels to dynamically optimize the temporal scale to generate the action instances.
However, these works cannot flexibly localize action instances. Unlike the previous methods, the anchor-free mechanism~\cite{10.1007/978-3-030-01225-0_1,8237579,7780583,8099825} did not rely on anchors, so it could flexibly predict the boundaries of actions. 
For instance, Lin \emph{et al.}~\cite{10.1007/978-3-030-01225-0_1} proposed the Boundary-Sensitive Network (BSN) based on local to global features to generate high-quality action proposals. Zhao \emph{et al.}~\cite{8237579} proposed the structured segment network (SSN) to model the temporal structure by a structured temporal pyramid. 
In recent years, balancing the previous two mechanisms~\cite{liu2019multi,9298475} has aroused the interest of researchers. Liu \emph{et al.}~\cite{liu2019multi} proposed a multi-granularity generator (MGG) to generate proposals from different granularities. Su \emph{et al.}~\cite{9298475} proposed a progressive cross-granularity cooperation (PCG-TAL) framework to effectively take advantage of complementary between the anchor-based and the frame-based ones. These methods all relied on accurate time annotations, leading to heavy annotation burden.

\subsection{Weakly-supervised temporal action localization}
WTAL solves the same problem but only refers to the video-level labels. Current works regarded WTAL as the Multiple Instance Learning (MIL)~\cite{cheplygina2019not} problem or utilized the attention mechanisms~\cite{8953929}. Wang \emph{et al.}~\cite{wang2017untrimmednets} presented a framework called UntrimmedNet which contained a classification module and a selection module. Paul \emph{et al.}~\cite{paul2018w} proposed the co-activity similarity loss based on paired videos of the same category to aggregate the feature similarity. Shou \emph{et al.}~\cite{shou2018autoloc} proposed an Outer-Inner-Contrastive (OIC) loss to discover the boundaries of actions rather than using thresholds. 
Islam \emph{et al.}~\cite{9093620} proposed a classification module and a deep metric learning module. The former calculated video-level predictions and the latter measured the similarity between different action instances. 

Recent works have gradually focused on the causes of poor WTAL performance. Some works~\cite{8953341,yu2019temporal,8737877} were devoted to solving the problem of action incompleteness. 
Liu \emph{et al.}\cite{8953341} proposed a multi-branch framework to model the completeness of actions based on the Diversity Loss. 
Yu \emph{et al.}~\cite{yu2019temporal} explicitly modeled the temporal structure, i.e. the start, the middle and the end, thereby effectively mining the potential temporal structure of action instances. 
Zeng \emph{et al.}~\cite{8737877} proposed an iterative-winners-out network, which selected the most discriminative action instances and removed them in the next training iteration. 
Meanwhile, other efforts~\cite{8737877,9157522} were made to alleviate the background false positive problem. 
Lee \emph{et al.}~\cite{lee2020background} proposed a framework called BaSNet, which consisted of a base branch and a suppression branch, to suppress the background to improve the performance. 
Shi \emph{et al.}~\cite{9157522} proposed to utilized Conditional Variational Autoencoder (VAE) to model the class-agnostic frame-wise probability to distinguish action frames from background frames. 
Although the video-level annotation can relieve the burden of labeling, the performance of the models trained by limited supervision are not comparable with the fully-supervised counterpart.

\subsection{Point supervision}
Regarding to the trade-off between the full supervision and the weak supervision, point supervision~\cite{bearman2016s,laradji2018blobs,laradji2019instance} has gradually attracted the attention of the community. Point supervision has been widely studied in the image domain where the annotation is a single-pixel label per object instance. Bearman \emph{et al.}~\cite{bearman2016s} applied point supervision to image semantic segmentation. Laradji \emph{et al.}~\cite{laradji2018blobs,laradji2019instance} utilized the point supervision for object counting and instance segmentation. Recently, point supervision has been introduced to the video domain as single-frame supervision. Moltisanti \emph{et al.}~\cite{8954471} proposed to classify the actions based on single-frame supervision. Ma \emph{et al.}~\cite{10.1007/978-3-030-58548-8_25} first introduced single-frame supervision into TAL task and proposed a framework called SF-Net. Our method is different from SF-Net~\cite{10.1007/978-3-030-58548-8_25} in two main aspects. (1) We propose a method to mine the pseudo snippet-level ground truth with the help of the Dilation-Erosion module. (2) Our method is able to mine the evident background as well as the hard background, while SF-Net~\cite{10.1007/978-3-030-58548-8_25} only focuses on the former.

\section{Method}
\label{method}

In this section, we first formalize the problem of Single-Frame Supervised Temporal Action Localization (SF-TAL). Then we introduce the details of our framework and finally give the details of the training and inference phases.

\textbf{Problem Formulation.}
Given a training video $\bm{v}_n\in\{\bm{v}_1,…,\bm{v}_N\}$, we adopt $\bm{y}_n\in\{0,1\}^{N_c}$ to denote the video-level label, where $N$ is the number of videos and $N_c$ is the number of action classes. Single-frame labels of video $v_n$ are denoted by $\{(t_l,\bm{c}_l)\}_{l=1}^M$, indicating that an action instance with category $\bm{c}_l\in\{0,1\}^{N_c+1}$ occurs at timestamp $t_l$. $M$ is the number of action instances in video $v_n$, 
and the $(N_c+1)^{\mathrm{th}}$ class represents the background class, denoted by $\bm{c}_b$. 
Note that $\bm{y}$ (the video index $n$ is omitted for simplicity) is a multi-hot vector, $\bm{c}_l$ and $\bm{c}_b$ are one-hot vectors. The goal of the TAL task is to generate a set of action segments, each of which is denoted by $(s ,e ,c ,q)$, for a test video. Here, $s$, $e$, $c$ denote the start time, the end time, and the category of the detection respectively, and $q$ represents its confidence score.

\begin{figure}[ht]
\centering
\includegraphics[scale=0.3]{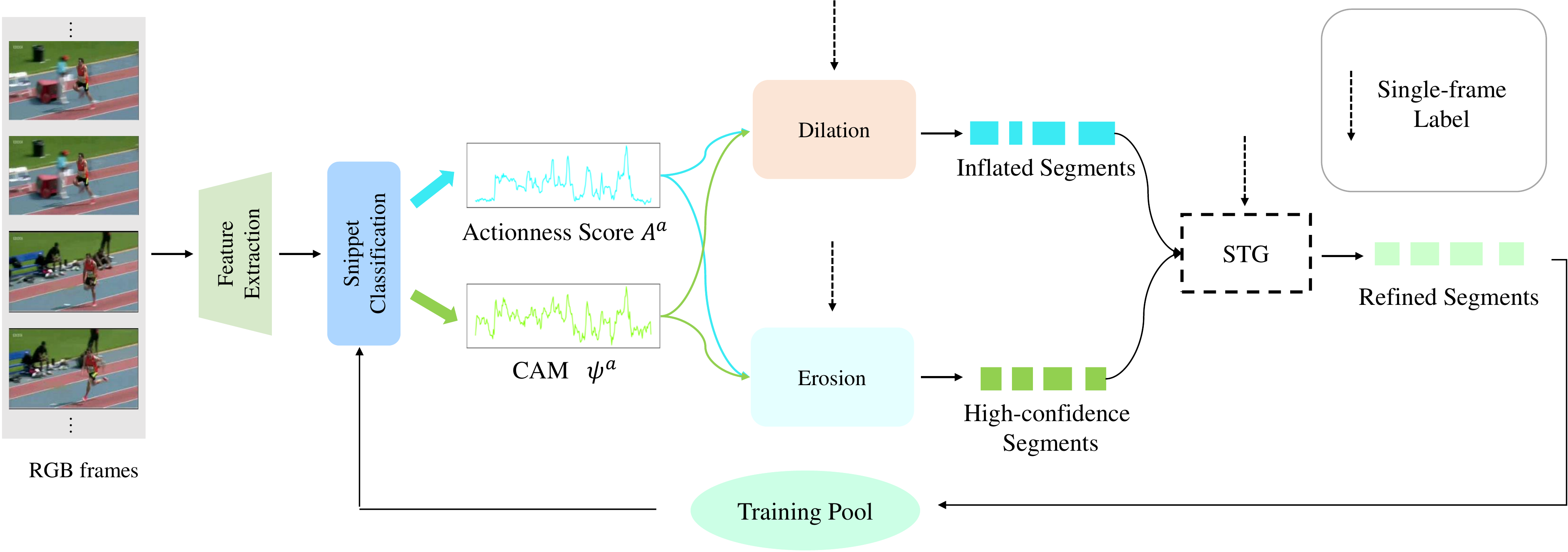}
\caption{Illustration of the architecture of the proposed method, and the framework consists of three parts: (1) Feature Extraction, (2) Snippet Classification, and (3) Dilation-Erosion. 
The main purpose of the Snippet Classification module is to predict actionness score $\bm{A}^\mathrm{a}$ and CAMs $\bm{\Psi}^\mathrm{a}$. 
The function of the Dilation-Erosion module is to mine the pseudo snippet-level ground truth, hard backgrounds and evident backgrounds based on $\bm{A}^\mathrm{a}$, $\bm{\Psi}^\mathrm{a}$, and single-frame annotations. 
Specifically, we first extract the coarse segments based on the class activate map (CAM), then apply late-fusion and early-fusion on CAM and the actionness score to obtain inflated segments and high-confidence segments. Finally, we use Seed Temporal Growing (STG) algorithm to extract refined segments and the background is not shown. 
The pseudo snippet-level ground truth is further used as supervision to train the network. The two modules form a cyclic dependency. In the end, we use the fused CAMs of RGB and optical flow to predict.}
\label{fig:framework}
\end{figure}

\subsection{Overview} 

Our proposed framework processes the appearance (RGB) and motion (optical flow) in two parallel streams, which are fused at the last stage. 
As the structure of the two streams are the same, we illustrate it with RGB stream (See Fig. \ref{fig:framework}).
Specifically, the input video is firstly processed by the Feature Extraction module to obtain high-level features which are fed into the Snippet Classification module. Then the outputs of the Snippet Classification module of both the streams are fused, which are used for inference. Importantly, during the training phase, the Dilation-Erosion module relies on the output of the Snippet Classification module to mine the pseudo snippet-level ground truth by which the Snippet Classification module is further trained, i.e., the two modules form a cyclic dependency. We describe the details of each module in the following sections.

\subsection{Feature Extraction}

We first extract the RGB frames and the optical flow from the video and split them into fixed-length and non-overlapping snippets, which are fed into the feature extraction network to obtain the snippet-wise feature $\bm{X}^\mathrm{a}\in\mathbb{R}^{T\times D}$ and $\bm{X}^\mathrm{f}\in\mathbb{R}^{T\times D}$, where $T$ is the number of snippets and $D$ is the feature dimension. Note that the superscript $\mathit{a}$ denotes the feature belonging to RGB, and the superscript $\mathit{f} $ denotes the feature belonging to optical flow. For simplicity, we only use the superscript $\mathit{a}$ (RGB) to illustrate. Following~\cite{8953341,9008791,lee2020background}, we use the I3D~ \cite{carreira2017quo} which has been pre-trained on Kinetics~\cite{carreira2017quo} as our feature extraction network.

\begin{figure}[ht]
\centering
\includegraphics[scale=0.45]{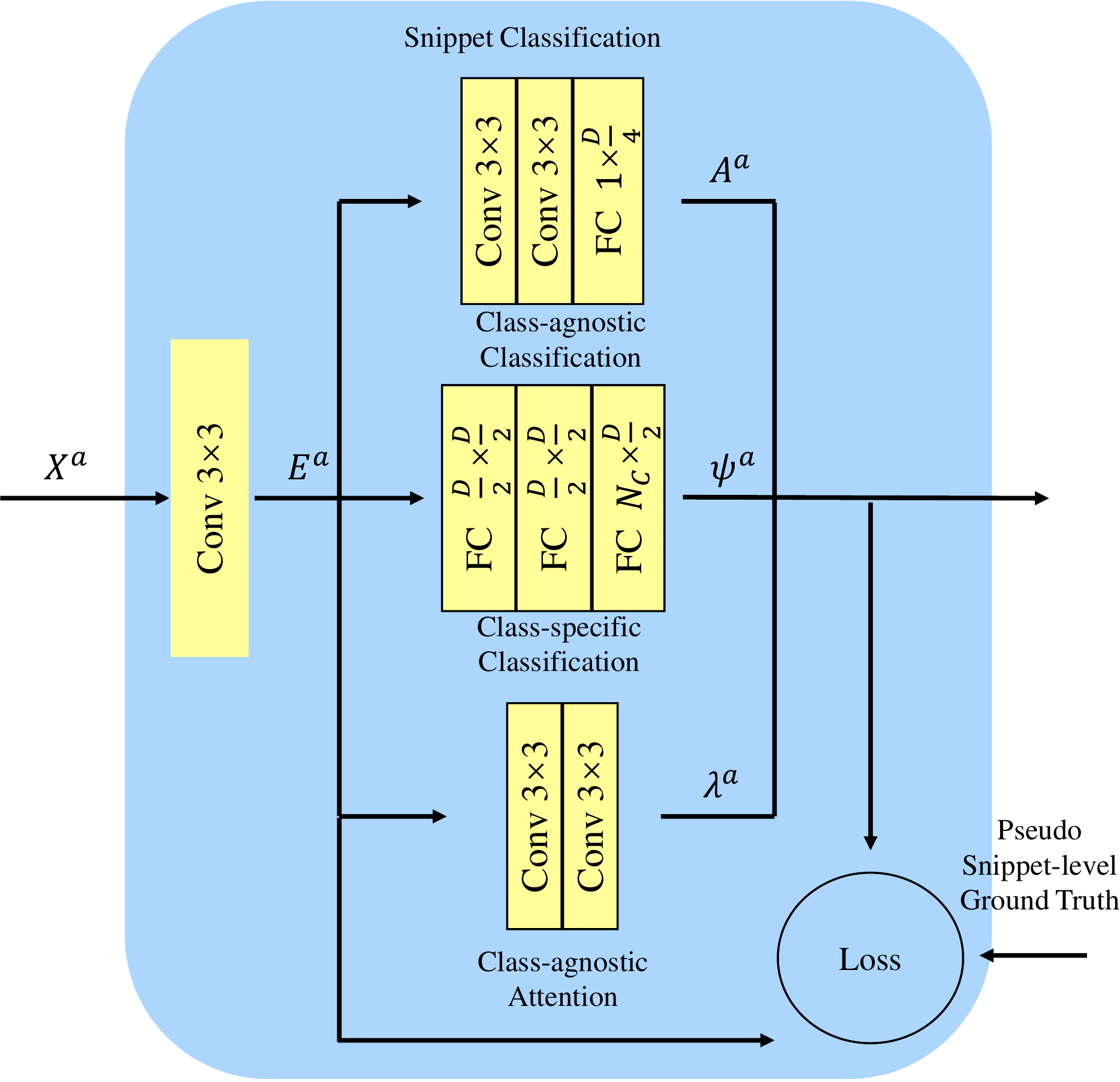}
\caption{
    Illustration of the architecture of the Snippet Classification module, which consists of four sub-modules: 
    feature embedding, class-agnostic classification, class-specific classification, and class-agnostic attention.
    The Conv $3 \times 3$ means the kernel size of the 1D Convolutional layer is 3.
    The FC $N_C \times \frac{D}{2}$ denotes the size of input and output of the fully-connected layer are $\frac{D}{2}$ and $N_C$, respectively.
}
\label{fig:SP}
\end{figure}

\subsection{Snippet Classification}
The module consists of four sub-modules: Feature Embedding, Class-agnostic Classification, Class-specific Classification, and Class-agnostic Attention. 
Next, we will introduce the details of each sub-module.

\emph{1) Feature Embedding.} To embed the feature $\bm{X}^\mathrm{a}$ for the adaptation to the task of temporal action localization, we utilize a temporal 1D convolutional layer~\cite{8953341} to obtain the embedded feature $\bm{E}^\mathrm{a}$. The embedded feature then flows into three following modules, as shown in Fig \ref{fig:SP}. 

\emph{2) Class-agnostic Classification.} The goal of class-agnostic classification is to obtain the binary actionness score $\bm{A}^\mathrm{a}\in\mathbb{R}^T$ which evaluates the probability of action occurrence in each snippet. Similar to~\cite{10.1007/978-3-030-58548-8_25,8237579,10.1007/978-3-030-01225-0_1}, we adopt two temporal 1D convolutional layers and one fully-connected layer.

\emph{3) Class-specific Classification.} Class-specific classification generates the  Class Activate Map (CAM) $\bm{\Psi}^\mathrm{a}\in\mathbb{R}^{T\times (N_c+1)}$, which represents the probability of each snippet for each action class (including background).  Specifically, we adopt three fully-connected layers to predict CAM.

\emph{4) Class-agnostic Attention.} To generate the class-agnostic attention score $\bm{\lambda}^\mathrm{a}\in\mathbb{R}^T$, which denotes the importance of each snippet,  we adopt two temporal 1D convolutional layers \cite{8953341,8954471}. It is worth mentioning that the sub-module of class-agnostic classification is trained with the snippet-level labels (from the training pool) to evaluate the actionness probability of each snippet, while the sub-module of class-agnostic attention is trained with the video-level label to evaluate the significance of each snippet in the video-level prediction.

\textbf{Fusion.} 
In this module, we fuse the outputs of the snippet classification of the RGB and the optical flow streams. Specifically, we concatenate the embedded features $\bm{E}^\mathrm{a}$ and $\bm{E}^\mathrm{f}$ into the fused embedded feature $\bm{E}^\mathrm{F}=[\bm{E}^\mathrm{a},\bm{E}^\mathrm{f}]$. Following \cite{9008791}, the actionness scores, the class activate maps, and class-agnostic attention scores are weighted by the learned parameters, and later combined by addition as the fused results ($\bm{A}^\mathrm{F}, \bm{\Psi}^\mathrm{F}, \bm{\lambda}^\mathrm{F}$). The Fused-CAM $\bm{\Psi}^\mathrm{F}$ will be used for inference as explained in Section \ref{inference}.

\subsection{Dilation-Erosion} To take full advantage of single-frame annotations to address the two challenges, \emph{action incompleteness} and \emph{background false positives}, we propose the Dilation-Erosion module including Segment Dilation and Segment Erosion. 
\textbf{For the first challenge}, we adopt the Segment Dilation to explore sufficiently large potential action segments to ensure the integrity of actions. \textbf{For the second challenge}, we perform the Segment Erosion to remove backgrounds in the potential action segments to suppress background false positives.
With the help of the Dilation-Erosion module, we obtain the pseudo snippet-level ground-truth, hard backgrounds and evident backgrounds.
Next, we present the details of the Dilation-Erosion module.

\textbf{Segment Dilation.} To alleviate the problem of \emph{action incompleteness}, we propose the Segment Dilation step to mine as many action segments as possible. Coarse segments are extracted based on CAM and have high response on the labeled category, while auxiliary segments are extracted based on the actionness score and have high probability of action occurrence. 
Inflated segments are determined by combining the two kinds of segments in a late-fusion style, and by considering the labeled frames as well.

\emph{1) Coarse segments.} Firstly, we select the CAM $\bm{\Psi}^{\mathrm{a}}_l\in\mathbb{R}^T$ from the CAMs $\bm{\Psi}^\mathrm{a}$ with the annotated category $\bm{c}_l$ to extract the coarse segments, which are above the threshold defined as follows,
\begin{align}
\tau=min(\bm{\Psi}^{\mathrm{a}}_l)+(max(\bm{\Psi}^{\mathrm{a}}_l)-min(\bm{\Psi}^{\mathrm{a}}_l))\times \eta,
\end{align}
where $\eta$ is a hyperparameter, determining the lengths of the coarse segments.

\emph{2) Inflated segments.} We obtain the inflated segments by combining CAM and the actionness score in a late-fusion style. Specifically, we obtain the auxiliary segments based on the actionness score $\bm{A}^\mathrm{a}$ and the threshold defined as  $\tau^\mathrm{a}=min(\bm{A}^\mathrm{a})+(max(\bm{A}^\mathrm{a})-min(\bm{A}^\mathrm{a}))\times \eta $. 
Inflated segments are defined by taking the union of the coarse segments and the auxiliary segments, which contain the same single-frame annotations.

\textbf{Segment Erosion.} In the Segment Erosion step, we aim to suppress background false positives by removing the backgrounds from the inflated segments. Specifically, we first extract high-confidence segments from the coarse segments. They will be used as the basis of determination of ``Erosion''. 
After this step, we obtain the refined segments with the STG algorithm (see Algorithm \ref{STG}).

\emph{1) High-confidence segments.} High-confidence segments are extracted from the coarse segments based on the assumption that the snippets with high confidence scores around the labeled snippet are highly likely to be in the action instance. Here, we define a corrected score as the confidence score, by fusing CAM and the actionness score in an early-fusion style as
\begin{align}
\hat {\bm{\Psi}}^{\mathrm{a}}_l=( \bm{\Psi}^{\mathrm{a}}_l+\bm{A}^{\mathrm a}).
\end{align}
Then a high-confidence segment is defined as a continuous snippet series including a labeled snippet, with the confidence scores greater than the median of those in a coarse segment.

\begin{algorithm}
\caption{Seed Temporal Growing (STG)}  
\label{STG}
\renewcommand{\algorithmicrequire}{\textbf{Input:}}
\renewcommand{\algorithmicensure}{\textbf{Output:}}
  \begin{algorithmic}
		\Require
		\\
        inflated segment $\bm{I}$, high-confidence segment $\bm{H}$, labeled snippet $s$;  
		\Ensure
		\\
        start time $first$, end time $last$;
  \end{algorithmic} 
  \begin{algorithmic}[1] 
        \State{Compute the evaluation sequences $\bm{F}^\mathrm{I}$, $\bm{F}^\mathrm{H}$ of $\bm{I}$, $\bm{H}$ using Eq. \ref{con:eva}\;}
        \State{Compute the threshold $\epsilon={median}(\bm{F}^\mathrm{H})$\;}
		\State{\textbf{for} $i \leftarrow s$ \textbf{downto} $0$ \textbf{do}}
        \State{\indent \textbf{if} $\bm{F}^\mathrm{I}[i] \ge \epsilon$ \textbf{then}}
		\State {\indent \indent $first = i$\;}
		\State {\indent \textbf{end}}
		\State {\textbf{end}}
		\State{\textbf{for} $i \leftarrow s$ \textbf{to} $len(\bm{I})$ \textbf{do}}
        \State{\indent \textbf{if} $\bm{F}^\mathrm{I}[i] \ge \epsilon$}
		\State {\indent \indent $last = i$\;}
		\State {\indent \textbf{end}}
		\State {\textbf{end}}
        \State{return $first, last$\;}
  \end{algorithmic}
\end{algorithm}

\emph{2) Refined segments.} To locate the boundaries of actions in the inflated segments, inspired by the Seed Region Growing algorithm \cite{huang2018weakly}, we propose a simple but effective algorithm named Seed Temporal Growing (STG) as introduced in Algorithm \ref{STG}. Specifically, the process starts from the labeled snippets and grows to the two directions along the temporal axis. Each snippet in an inflated segment is determined if it should be included in the action segment successively. If a snippet satisfies the condition, the boundary is updated; otherwise, the growing process in this direction continues, but the boundary is not updated. 
In a word, the STG algorithm returns the leftmost and rightmost snippets that satisfy the condition.

Apparently, a key factor of the STG algorithm is the condition. Here, we propose to evaluate the $t^{\mathrm{th}}$ snippet based on its corrected score and its feature similarity with the labeled snippet by the function
\begin{align}
F(t)=\hat {\bm{\Psi}}^{\mathrm{a}}_{l}(t)\times e^{-d(\bm{X}_t^\mathrm{a},\bm{X}_l^\mathrm{a})},t\neq l\label{con:eva},
\end{align}
where $d(\cdot)$ is the cosine similarity and $l$ denotes the labeled snippet. The median of the evaluation score $\bm{F}^\mathrm{H}$ of the high-confidence segment  $\bm{H}$ inside the inflated segment $\bm{I}$ is adopted as the threshold $\epsilon$. 
Finally, the parts between the two boundaries are considered as refined segments, as well as action instances for the pseudo ground truth.

\emph{Background.} To suppress the background false positives, we mine two types of background, i.e., the hard background and the evident background. The former is the main cause of background false positives. Here, we define it as the coarse segment which has no intersection with any refined segment as,

\begin{align}
IoU(\bm{p}_i,\bm{p}_j) = 0
\label{iou}
\end{align}
where $\bm{p}_i$ is a coarse segment and $\bm{p}_j$ is a refined segment. 

For the evident background, following \cite{10.1007/978-3-030-58548-8_25}, we compute the background score as $\bm{\hat b} =(-\bm{A}^\mathrm{a}+\bm{b})/2$, where $\bm{b}\in\mathbb{R}^T$ is the CAM of the background category from the CAMs $\bm{\Psi}^\mathrm{a}$. Then the snippets with the top-k background score ( $k=\lceil T/8 \rceil$ ) are labeled as evident backgrounds.

\begin{figure}[ht]
\centering
\includegraphics[scale=0.45]{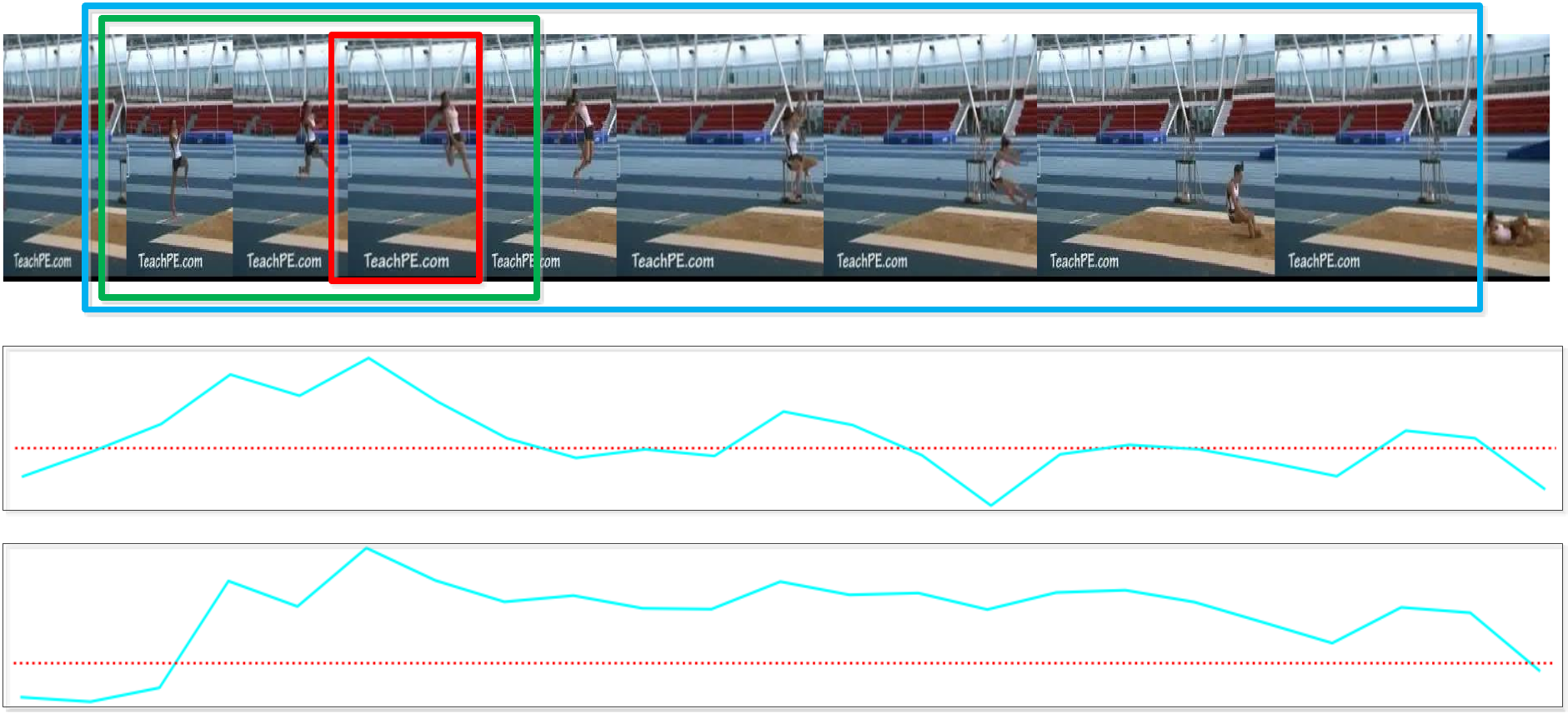}
\caption{Visualization of finding the refined segment and the CAM in different training stages. \textbf{Top}: The red bounding box represents the labeled frame, the green bounding box represents the high-confidence segment, and the blue bounding box represents the refined segment. \textbf{Middle}: the curve represents the CAM obtained after the first training stage, and the red line represents the threshold. \textbf{Bottom}: the curve represents the CAM of the model retained by the pseudo snippet-level ground truth. We can see that less discriminative parts of the action instance can also be detected.}
\label{fig:action}
\end{figure}

After the Dilation-Erosion module, we obtain the pseudo snippet-level ground-truth, hard backgrounds and evident backgrounds and put them into the training pool to retrain the network, thereby effectively making the actions more complete and reducing the background false positives. To better illustrate the process and its benefit to the performance of the model, we present an example in Fig. \ref{fig:action}. It can be seen that the model performs better after being retrained under the supervision of the pseudo snippet-level ground-truth generated by the Dilation-Erosion, as the less discriminative parts of the action can also be detected. The training strategies will be introduced in the following section.

\subsection{Training}
\label{sec:loss}

\noindent Our framework is trained using the overall loss formulation
\begin{align}
\mathcal{L} = {\mathcal L}_{\mathrm{cls\_video}} + {\mathcal L}_{\mathrm{cls\_snippet}} + {\mathcal L}_{\mathrm{action}} + {\mathcal L}_{\mathrm{emb}},
\end{align}
where ${\mathcal L}_{\mathrm{cls\_video}}$, ${\mathcal L}_{\mathrm{cls\_snippet}}$, ${\mathcal L}_{\mathrm{action}}$ and ${\mathcal L}_{\mathrm{emb}}$ all include three parts: the appearance (RGB), the optical flow and the fusion. 
We train the framework in Fig. \ref{fig:framework} with RGB and optical flow, and then the fusion of them. 
Next, we describe the details of the loss terms (Take RGB for example).

\emph{1) Video classification loss.} Following~\cite{9008791}, the probability distribution of the video-level action categories is defined as

\begin{align}
\bm{p}^\mathrm{a}=softmax(\sum_{t=1}^{T}\bm{\lambda} _t^\mathrm{a}\times \bm{\Psi}_t^{\mathrm{a}}),
\label{snippet-level}
\end{align}
where $\bm{\lambda} _t^\mathrm{a}$ and $\bm{\Psi}_t$ are the class-agnostic attention score and CAM of the $t^{\mathrm{th}}$ snippet. Here, the prediction of the background class is not considered. Then the video classification loss is defined as
\begin{align}
\mathcal{L}_{\mathrm{cls\_video}}^\mathrm{a}=-\sum_{c=1}^{N_c}\bm{y}^c\log \bm{p}_c^\mathrm{a}.
\end{align}

\emph{2) Snippet classification loss.} To predict the score of each snippet for each action class, following~\cite{10.1007/978-3-030-58548-8_25}, we define the loss function as
\begin{equation} 
\begin{split} 
\mathcal{L}_{\mathrm{cls\_snippet}}^\mathrm{a}=\sum_{l=1}^{M}(-\frac{1}{N_l}\sum_{i=1}^{N_l} \bm{c}_l&\log \bm{\Psi}^{\mathrm{a}}_{i,l})\\
&-\mu\frac{1}{N_b}\sum_{j=1}^{N_b} \bm{c}_\mathrm{b}\log \bm{\Psi}^{\mathrm{a}}_{j,b}.\\
\end{split} 
\end{equation}

\noindent Here, for the $l^{\mathrm{th}}$ annotated frame, $\bm{c}_l$ is the action class, $N_l$ is the number of action snippets and $\bm{\Psi}^{\mathrm{a}}_{i,l}$ denote the CAM of the $i^{\mathrm{th}}$ action snippet.
$c_b$ is the background class, $N_b$ is the number of background snippets and $\bm{\Psi}^{\mathrm{a}}_{j,b}$ denote the CAM of the $j^{\mathrm{th}}$ background snippet.%

\emph{3) Actionness loss.} To evaluate the actionness probability of each snippet, inspired by~\cite{10.1007/978-3-030-01225-0_1}, we define the actionness loss $\mathcal{L}_{\mathrm{action}}^\mathrm{a}$ as

\begin{equation} 
\begin{split} 
\mathcal{L}_{\mathrm{action}}^\mathrm{a}=\sum_{l=1}^{M}(-\frac{1}{N_l}\sum_{i=1}^{N_l}& \log (\bm{A}^{\mathrm{a}}_{i,l}))\\
&-\mu \frac{1}{N_b}\sum_{j=1}^{N_l} \log (1-\bm{A}^{\mathrm{a}}_{j,b}),\\
\end{split} 
\end{equation}
where $\bm{A}^{\mathrm{a}}_{i,l}$ and $\bm{A}^{\mathrm{a}}_{i,b}$ are the actionness scores of the $i^{\mathrm{th}}$ action snippet for the $l^{\mathrm{th}}$ annotated frame and $j^{\mathrm{th}}$ background snippet.

\emph{4) Embedding loss.} To aggregate the features of action instances with the same category and to separate the features of actions from backgrounds, similar to \cite{laradji2019instance}, we construct the snippet pair set $P$ which contains action-action and action-background pairs. Firstly, the action set is defined by the refined segment set and the background set is defined by the evident background set. Then snippet pairs are randomly selected to form the snippet pair set $P$. The embedding loss is defined as
\begin{equation} 
\begin{split} 
{\mathcal L}_{\mathrm{emb}}^\mathrm{a}=-\frac{1}{N_p} \sum_{(i,j)\in P}& [\mathbbm 1 _{\{\bm{c}_i=\bm{c}_j\}}\log (e^{-d(\bm{E}_i^\mathrm{a},\bm{E}_j^\mathrm{a})}) \\
           &+\mathbbm 1 _{\{\bm{c}_i\neq \bm{c}_j\}}\log (1-e^{-d(\bm{E}_i^\mathrm{a},\bm{E}_j^\mathrm{a})})],	 
\end{split} 
\end{equation}

\noindent where $\mathbbm{1}$ is the indicator function, $\bm{E}_i^\mathrm{a}$ and $\bm{E}_j^\mathrm{a}$ are the embedded features, and $\bm{c}_i$, $\bm{c}_j$ are the snippet-level labels.

Overall, the proposed framework contains two stages during the training phase. 
In the first stage, we train the framework on the single-frame supervision. 
To minimize the impact of initialization, we select $u$ adjacent snippets around the labeled snippet to train a coarse snippet classifier, and $u$ is randomly selected from the set $\{0,...,U\}$. 
According to \cite{10.1007/978-3-030-58548-8_25}, the annotators prefer to label frames near to the middle part of action instances. 
The number of snippets located in the left (or right) of the labeled snippet within the same action instance, is less than the half average duration of action instances in a dataset. 
As a result, the value of U is equal to the half average duration of action instances in a dataset.
No background sample is considered in the first stage. Therefore, the background part of ${\mathcal L}_{\mathrm{cls\_snippet}}$,  ${\mathcal L}_{\mathrm{action}}$, and ${\mathcal L}_{\mathrm{emb}}$ can be ignored. In the second stage, based on the coarse snippet classifier and the pseudo ground-truth generated by the dilation-erosion process, we re-train the network to obtain a refined snippet classifier.

\subsection{Inference}
\label{inference}
During the inference phase, we use the Fused-CAM to generate action detection. Following 3C-Net~\cite{9008791}, we only keep the action categories with the average top-k scores above 0. 
Then the detection of these action categories is generated by using the threshold $\varepsilon$ on Fused-CAM. 
In our experiment, the detection with the highest confidence score is used for evaluation.
The confidence score of a detection is defined as the sum of its highest prediction score and the category score as defined in Eq. \ref{snippet-level}. 
Note that we discard the CAM of the background class in the inference phase.

\section{Experiments}
\label{exp}

\subsection{Datasets and Evaluation Metrics}
\textbf{Datasets.} To evaluate the proposed method, extensive experiments are conducted on two challenging untrimmed datasets: THUMOS14~\cite{idrees2017thumos} and ActivityNet 1.2~\cite{caba2015activitynet}.

\emph{1) THUMOS14}~\cite{idrees2017thumos} consists of 101 action categories, 1010 validation videos, and 1574 test videos. Among them, 413 videos have been annotated with the boundaries and action categories of the action instances, including 200 validation videos and 213 test videos, with a total of 20 action categories. The average duration of action instances is 4.5 seconds, and the average number of action instances in each video is about 15. Following \cite{8953341,9008791,paul2018w}, we use the validation videos for training and the test videos for evaluation. In this work, we utilize the single-frame annotation (including timestamp and action category) of each action instance in the training set as the supervision.

\emph{2) ActivityNet 1.2}~\cite{caba2015activitynet} consists of 100 action categories, 4819 training videos and 2383 validation videos. The average duration of action instances is 49.9 seconds and the average count of action instances in each video is 1.5. Nearly half of the action instances last longer than 50\% of the duration of a video. Because the ground-truths of test videos are not public, following \cite{8953341,9008791}, we use the training videos to train the model and utilize the validation videos for evaluation. Similarly, we use the single-frame annotation as the setting of supervision.

\textbf{Evaluation Metrics:} For the temporal action localization task, we follow the standard protocol of evaluation. On the two datasets, the mean Average Precision (mAP) of the temporal intersection over union (tIoU) at different thresholds will be calculated respectively. If the category of a detection is consistent with the ground-truth, and the temporal IoU is greater than the IoU threshold, it is considered as positive detection. Following~\cite{10.1007/978-3-030-58548-8_25}, on the THUMOS14 dataset, we report the performance of 0.1:0.1:0.7 and the average mAP is computed on 0.1:0.1:0.5. On the ActivityNet 1.2 dataset, we report the performance of IoU threshold at 0.5,0.7,0.9, as well as the average mAP on 0.5:0.05:0.95.

\subsection{Implementation Details}
The annotated frames on the two datasets are selected by uniform sampling in the ground-truth of each action instance in the training videos. 
Each snippet, which contains 16 frames, is the input to the Feature Extraction module.
In the training stage, we train the model alternately with the original features and the randomly cropped features. For the random cropping, the fixed length T is set to 750 and 200 on Thumos14 dataset and ActivityNet1.2 dataset respectively. For the feature extraction network, we use I3D network pre-trained on Kinetics dataset, and adopt the output of the Mixed\_5c layers followed by the spatio-temporal average pooling as the feature $\bm{X}^\mathrm{a}\in\mathbb{R}^{T\times1024}$ and $\bm{X}^\mathrm{f}\in\mathbb{R}^{T\times1024}$. For the flow stream, we apply the TV-L1 \cite{zach2007duality} algorithm to generate the optical flow. Following \cite{8953341,9008791}, we do not fine-tune the backbone network. We use Adam \cite{kingma2014adam} optimizer with $10^{-4}$ learning rate and 0.005 weight decay to optimize the network. Hyperparameters are determined empirically. $\mu$ is set to 0.1 for both the datasets. $\eta$ is set to 0.5 and 0.2, and $U$ is set to 2 and 25 for THUMOS14 and ActivityNet 1.2 datasets respectively. The threshold $ \varepsilon$ for localization is set to $mean(\varphi)$ ($\varphi$ is the class activate map) for THUMOS14 and set to 0 for ActivityNet 1.2.

\begin{table}[h!]
  \begin{center}
  \caption{Temporal action localization performance comparison (mAP) of the proposed approach with state-of-the-art methods on the THUMOS14 dataset. We report the mAP values at different IoU thresholds, and AVG represents the average value of the results when the IoU threshold is [0.1:0.1:0.5]. Our method outperforms existing weakly-supervised methods and achieves 5.1\% at IoU = 0.5, compared to the best weakly-supervised result \cite{2020Weakly}. Superscript `+’ indicates that there is not only video-level labels during training. `manually’ denotes single-frame labels are manually annotated by human annotators.`uniform' means that the single-frame labels are sampled on the groundtruth according to a uniform distribution.}
    \begin{tabular}{ccccccccc}
      \toprule
      \toprule
	 \multicolumn{1}{c|}{\multirow{2}{*}{Methods}}& \multicolumn{8}{c}{\multirow{1}{*}{mAP@IoU}} \\  
      \multicolumn{1}{c|}{} & 0.1 & 0.2 &0.3 &0.4 &0.5 &0.6 &0.7 &AVG\\ 
      \midrule
      \midrule
	 \multicolumn{9}{c}{\multirow{1}{*}{Fully-supervised Methods}} \\
      \midrule
	\multicolumn{1}{c|}{S-CNN\cite{shou2016temporal}}&47.7 &43.5 &36.3 &28.7 &19.0 &-  &5.3 &35.0 \\
	\multicolumn{1}{c|}{SSN\cite{8237579}}&60.3	&56.2	&50.6	&40.8 &29.1 &-	&- &47.4 \\
	\multicolumn{1}{c|}{BSN\cite{10.1007/978-3-030-01225-0_1}}&-	&-	&53.5 	&45.0 	&36.9 &28.4 	&20.0	 &- \\
	\multicolumn{1}{c|}{MGG\cite{liu2019multi}}&-	&-	&53.9	&46.8&	37.4	&29.5&	21.3	&- \\
	\multicolumn{1}{c|}{BoW-Network+TCN\cite{9021961}}&68.5	&63.6	&55.7	&45.0&	31.6	&-&	-	&52.9 \\
	\multicolumn{1}{c|}{GTAN\cite{long2019gaussian}}&69.1	&63.7	&57.8	&47.2	&38.8	&-	&-	&55.3 \\
	\multicolumn{1}{c|}{P-GCN\cite{zeng2019graph}}&69.5	&67.8	&63.6	&57.8	& 49.1	&-	&-	& 61.6 \\
	\multicolumn{1}{c|}{Zhao \emph{et al}.\cite{zhao2020bottom}}& -	& -	& 53.9	& 50.7	& 45.4	& 38.0	& 28.5	& - \\
	\multicolumn{1}{c|}{SALAD \cite{vaudaux2021salad}}& 73.3	& 70.7 & 65.7 & 57.0 &44.6&	-	&-	& 62.3 \\
	\multicolumn{1}{c|}{AFSD\cite{lin2021learning}}& -	& -	& 67.3	& 62.4	& 55.5	& 43.7	& 31.1	& - \\
      \midrule
      \midrule
	 \multicolumn{9}{c}{\multirow{1}{*}{Weakly-supervised Methods}} \\
      \midrule
	\multicolumn{1}{c|}{W-TALC\cite{paul2018w}}&55.2&	49.6	&40.1&	31.1	&22.8	&-	&7.6	&39.8 \\
	\multicolumn{1}{c|}{Liu \emph{et al}.\cite{8953341}}&57.4	&50.8	&41.2	&32.1	&23.1	&15.0	&7.0	&40.9 \\
	\multicolumn{1}{c|}{3C-Net$^+$\cite{9008791}}&59.1&	53.5	&44.2	&34.1	&26.6&	8.1	&-	&43.5 \\
	\multicolumn{1}{c|}{BaSNet\cite{lee2020background}}&58.2&	52.3	&47.6	&36.0	&27.0	&18.6	&10.4&	43.6 \\
	\multicolumn{1}{c|}{DGAM\cite{9157522}}&60.0&	54.2	&46.8	&38.2	&28.8	&19.8	&11.4&	45.6 \\
	\multicolumn{1}{c|}{Maheen \emph{et al}.\cite{9093404}}&63.7&	56.9	&47.3	&36.4	&26.1	&-	&-&	46.1 \\
	\multicolumn{1}{c|}{Ding \emph{et al}.$^+$\cite{2020Weakly}}&61.6	&55.8&	48.2	&39.7	&31.6&	22.0	& 13.8	&47.4 \\
	\multicolumn{1}{c|}{SODA$^+$\cite{zhao2021soda}}&-&-& 53.1	& 44.9	& 35.6	&26.4 & 15.8 &-\\
      \midrule
	\multicolumn{1}{c|}{SF-Net$^+$\cite{10.1007/978-3-030-58548-8_25} manually}&\bf71.0	&63.4&	53.2	&40.7	&29.3&	18.4	&9.6	&51.5 \\
	\multicolumn{1}{c|}{Ours$^+$ manually}&68.6 &63.3 &\bf55.6 &\bf47.3 &\bf37.6 &24.4& 13.0&\bf 54.5 \\
      \midrule
	\multicolumn{1}{c|}{SF-Net$^+$\cite{10.1007/978-3-030-58548-8_25} uniform}&68.3	&62.3&	52.8	&42.2	&30.5&	20.6	&12.0	&51.2 \\
	\multicolumn{1}{c|}{Ours$^+$ uniform}&68.8 &\bf63.7 &54.0 &45.4 &36.3 &\bf25.0& 13.5& 53.6 \\
      \bottomrule
      \bottomrule
    \end{tabular}
    
    \label{thu}
  \end{center}
\end{table}

\subsection{Comparisons with the State-of-the-Art}
In this section, the proposed single-frame TAL method is compared with the state-of-the-art weakly supervised and fully supervised TAL methods. 
The experimental results on THUMOS14 dataset are shown in Table \ref{thu}. 
Superscript `+' indicates that there is extra kind of annotation besides the video-level labels. 
3C-Net~\cite{9008791} adds the count information of each action category. 
Ding \emph{et al.}~\cite{2020Weakly} adds segment annotation (two timestamps + action category) of each action instance. 
SODA~\cite{zhao2021soda} adds the single-action-frame and the single-background-frame for each action instance.
Among the existing methods, the SF-Net~\cite{10.1007/978-3-030-58548-8_25} is the only one under the same supervision as ours.

Our method outperforms the existing weakly supervised methods at most IoU thresholds as expected. There is a gain of 8\% in terms of mAP at IoU=0.5 and a gain of 7\% of the average mAP compared with the traditional weakly-supervised methods. We notice that the proposed method achieves comparable results although the annotation is much weaker than that of the fully supervised methods. It even surpasses many existing fully-supervised methods, and there is only a gap of 2\% of average mAP compared with GTAN~\cite{long2019gaussian}. 
\begin{figure}[ht]
\centering
\includegraphics[scale=0.9]{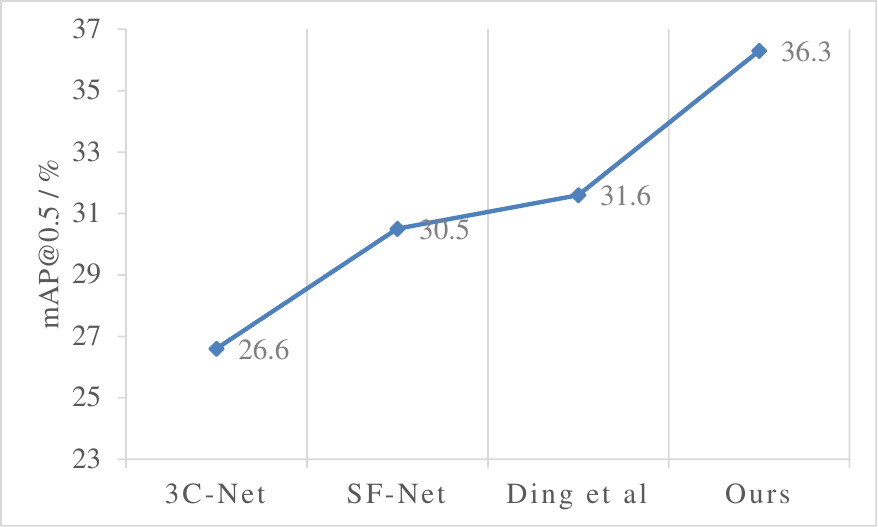}
\caption{MAP@0.5 on THUMOS14 dataset under different supervision.}
\label{fig:comp}
\end{figure}
It is observed that our method performs better than SF-Net~\cite{10.1007/978-3-030-58548-8_25} in both manual labeling and uniform labeling. Specifically, the mAPs at IoU=0.5 are improved by 8.3\% and 5.8\%, respectively. In addition, the mAP of our method is 4.7\% (IoU=0.5) higher than Ding \emph{et al.}~\cite{2020Weakly} (segment annotation).
It proves the effectiveness of the Dilation-Erosion strategy. However, it is also noticed that there are smaller improvements of mAPs at low IoUs, i.e. 0.1 - 0.3, compared with those at high IoUs. We believe it is because that our work focuses on relieving the problems of action incompleteness and false positives, which mainly benefit accuracy but not recall. In addition, we illustrate the comparison with several methods in Fig. \ref{fig:comp}, which shows that the performance generally improves as the supervision level increases.

Table \ref{act} lists the state-of-the-art comparison on ActivityNet 1.2 dataset. 
The proposed method is compared with the weakly supervised TAL as well as the fully supervised TAL. 
It can be seen that our method has achieved better performance compared with some weakly supervised methods. 
It is worth noting that, compared with THUMOS14,  the average count of action instances is relatively smaller, and the durations of action instances are quite longer. 
This weakens the impact of the Dilation-Erosion module for mining evident background and the hard background, resulting in less advantage of our framework to other methods.

\begin{table}[h!]
  \begin{center}
  \caption{Temporal action localization performance comparison (mAP) of the proposed approach with state-of-the-art methods on ActivityNet1.2 dataset. We report the mAP values at different IoU thresholds, and AVG represents the average value of the results when the IoU threshold is [0.5:0.05:0.95]. Superscript `+’ indicates that there is not only video-level labels during training.}
    \begin{tabular}{ccccc}
      \toprule
      \toprule
	 \multicolumn{1}{c|}{\multirow{2}{*}{Methods}}& \multicolumn{4}{c}{\multirow{1}{*}{mAP@IoU}} \\  
      \multicolumn{1}{c|}{}&0.5 &0.7 &0.9 &AVG\\ 
      \bottomrule
      \toprule
	 \multicolumn{5}{c}{\multirow{1}{*}{Fully-supervised Methods}} \\
      \midrule
	\multicolumn{1}{c|}{SSN\cite{8237579}}&41.3	&30.4	&13.2	&26.6 \\
      \bottomrule
      \toprule
	 \multicolumn{5}{c}{\multirow{1}{*}{Weakly-supervised Methods}} \\
      \toprule
	\multicolumn{1}{c|}{Autoloc\cite{shou2018autoloc}}&27.3&	17.5	&6.8	&16.0 \\
	\multicolumn{1}{c|}{W-TALC\cite{paul2018w}}& 37.0	&14.6	&4.2	&18.0\\
	\multicolumn{1}{c|}{TSM\cite{yu2019temporal}}& 28.3	&18.9	&7.5	&17.1\\
	\multicolumn{1}{c|}{Islam \emph{et al}.\cite{9093620}}&35.2 &16.3 &- &- \\
	\multicolumn{1}{c|}{CleanNet\cite{liu2019weakly}}&37.1 &23.4 &9.2 &21.6 \\
	\multicolumn{1}{c|}{3C-Net\cite{9008791}}& 37.2	&23.7	&9.2	&21.7 \\
	\multicolumn{1}{c|}{DGAM\cite{9157522}}&41.0&	26.9	&10.8	&24.4 \\
	\multicolumn{1}{c|}{SODA$^+$\cite{zhao2021soda}}& 42.0	&30.7	&14.6	&27.4 \\
      \midrule
	\multicolumn{1}{c|}{SF-Net$^+$\cite{10.1007/978-3-030-58548-8_25} uniform}& 37.8 & 24.6	& 10.3	& 22.8 \\
	\multicolumn{1}{c|}{Ours$^+$ uniform}& 37.3	& 25.1	& 10.1	& 22.5 \\
      \bottomrule
      \bottomrule
    \end{tabular}

    \label{act}
  \end{center}
\end{table}

\subsection{Ablation Studies}
To analyze the contribution of each component in the proposed method, i.e. the different loss terms, the steps in the Dilation-Erosion module and the two kinds of backgrounds, we conduct a set of ablation study experiments.

\emph{(1) Loss Functions.} The results of temporal action localization performance with different loss terms on THUMOS14 dataset are shown in Table \ref{strategy}. 
The first run, which is also the baseline, is conducted by training the model with the single video classification loss, denoted as $\mathcal{L}_{\mathrm{cls\_video}}$. 
The second and the third run, denoted as $\mathcal{L}_\mathrm{{cls\_snippet}}$ and $\mathcal{L}_{\mathrm{action}}$ respectively, are conducted by adding the snippet classification loss and the actionness loss to the baseline successively. 
The fourth run, denoted as $\mathcal{L}_{\mathrm{emb}}$, is conducted by replacing the actionness loss with the embedding loss in the third run. 
The fifth run includes all the four loss terms. 
Note that when $\mathcal{L}_{\mathrm{action}}$ is not included in training phase, we remove the class-agnostic classification sub-module. 
We can see that the performance of the model is improved compared to the baseline at AVG, in turn, by 14.2\%, 15.3\%, 15.8\% and 16.6\%. 
When labeled snippets are involved in the training, the performance of the model can be significantly improved (+14.2\%). 
The result indicates that the single-frame annotation is of great benefit to the TAL problem. 
The results of the third and the fourth runs improve the performance further, and prove the importance of the actionness loss and the embedding loss. 
As analyzed earlier, the embedding loss can aggregate the features of action instances with the same label and separate the features of actions from backgrounds. 
In another word, with the help of the embedding loss, the actionness loss is supposed to obtain more favorable performance, which is in line with the results of the fifth run.

\begin{table}[t]
\centering
\caption{ Results(mAP) with different loss terms on THUMOS14.}
\resizebox{0.95\columnwidth}{!}{
\begin{tabular}{l|cccc|ccccc}
    \toprule
    \toprule
    \multirow{2}{*}{Loss terms} & \multirow{2}{*}{$L_{cls\_video}$} & \multirow{2}{*}{$L_{cls\_snippet}$} & \multirow{2}{*}{$L_{action}$} & \multirow{2}{*}{$L_{emb}$} 
    & \multicolumn{5}{c}{mAP@IOU} \\
       &  &  &  &  &  0.1  &  0.3  &  0.5  & 0.7 & AVG   \\
    \midrule
                \multicolumn{1}{c|}{Run 1} & \cmark & & & & 54.0 & 35.6 & 20.1  & 6.4  & 36.6 \\
                \multicolumn{1}{c|}{Run 2} & \cmark  & \cmark & & & 65.1  & 51.1  & 34.4 & 12.2 & 50.8 \\
                \multicolumn{1}{c|}{Run 3} & \cmark  & \cmark & \cmark &  &  66.3 & 51.4 & 35.8 & 13.7 & 51.9 \\
                \multicolumn{1}{c|}{Run 4} & \cmark  & \cmark &  & \cmark &  67.3 & 53.0 & 35.0 & 12.4 & 52.4 \\
                \multicolumn{1}{c|}{Run 5} & \cmark  & \cmark & \cmark & \cmark &  \textbf{67.9}  & \textbf{53.4}  & \textbf{36.7} & \textbf{13.8}  & \textbf{53.2} \\
    \bottomrule
    \bottomrule
\end{tabular}
}
\label{strategy}
\end{table}

\emph{(2) Dilation-Erosion.} Table \ref{abation} shows the results with different dilation-erosion strategies on the THUMOS14 dataset.

\begin{itemize}

\item\textbf{NoDE}: Only the labeled frames are used as the supervision. In other words, we do not perform the strategy of Dilation-Erosion. It is also the baseline of the proposed method.

\item\textbf{NoDilation}: We remove the Dilation stage, which means we do not obtain the inflated segments and use the coarse segments instead.

\item\textbf{NoErosion}: We remove the Erosion stage from the method and use the coarse segments to replace the refined segments.

\item\textbf{NoHCS}: We remove the step of mining high-confidence segments from the Erosion stage and use the coarse segments instead. 

\end{itemize}

\begin{table}
\centering
	\caption{ Results(mAP) of different dilation-erosion strategies on THUMOS14 dataset. We report the mAP values at different IoU thresholds, and AVG represents the average value of the results when the IOU threshold is [0.1:0.1:0.5].}
    \begin{tabular}{ccccccccc}
      \toprule
      \toprule
      \multicolumn{1}{c|}{Strategies}&@0.3 &@0.5 &@0.7 &@AVG\\ 
      \midrule
	\multicolumn{1}{c|}{NoDE}&48.0 &28.9 &10.6 &41.4  \\
	\multicolumn{1}{c|}{NoDilation}&51.3 &35.8 &13.2 &51.3  \\
	\multicolumn{1}{c|}{NoErosion}&51.7 &34.1 &12.2 &51.6  \\
	\multicolumn{1}{c|}{NoHCS}&50.8 &34.9 &12.5 &51.1  \\
	\multicolumn{1}{c|}{Proposed}&54.0 &36.3 &13.5 &53.6  \\
      \bottomrule
      \bottomrule
    \end{tabular}

  \label{abation}
\end{table}

According to Table \ref{abation}, we can draw the following conclusions. (1) NoDE: Without the Dilation-Erosion strategy, the performance of the model drops by 7.8\% and 11.8\% at mAP@0.5 and average respectively, which proves the effectiveness of the proposed strategy. (2) NoDilation: The performance drops at low IoUs are more significant than those at high IoU when the inflate segments are not extracted. It indicates that the later-fusion of CAM and the actionness score is important to extracting sufficiently large potential action segments. (3) NoErosion: The performance drops at high IoUs are more significant than those at low IoU when the Erosion step is not included. It proves that the Erosion step benefits more to the accuracy of detection. (4) NoHCS: The mAP at low IoU declines more compared with NoErosion, and it proves that using high-confidence segments as the standard of `Erosion' 
can effectively remove the potential background and thus help the STG algorithm to locate the boundaries of action instances.

\begin{figure}[ht]
\centering
\includegraphics[scale=0.7]{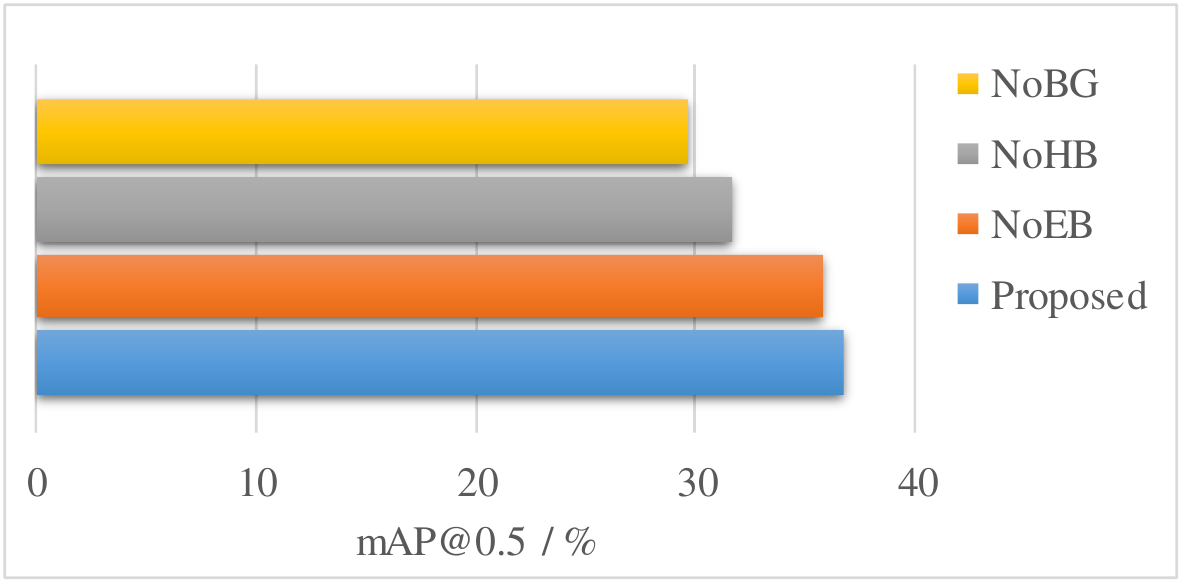}
\caption{Results (mAP) of mining different kinds of background on THUMOS14 dataset at IoU=0.5.}
\label{fig:ablation}
\end{figure}

\emph{(3) Background.} We conducte an ablation study to testify the effect of the two kinds of backgrounds on the overall performance. Fig \ref{fig:ablation} shows the results of four runs with mining different kinds of backgrounds on THUMOS14 dataset (at IoU=0.5).

\begin{itemize}
\item\textbf{NoBG}: without either of the backgrounds

\item\textbf{NoEB}: without the evident background.

\item\textbf{NoHB}: without the hard background.

\end{itemize}

From these experiment results, we can obtain the following conclusions. (1) NoBG: The performance is significantly worse (-7.1\% at IoU=0.5) than the proposed method, and it shows the importance of mining the two kinds of backgrounds to the system. (2) NoHB: Compared with the baseline (NoBG), the mAP@0.5 increases by 2\% when the evident backgrounds are mined. (3) NoEB: Compared with the baseline (NoBG), the mAP@0.5 increases by 6.2\% when the hard backgrounds are mined. It indicates that the hard background plays a more important role than the evident background for suppressing background false positives.

\begin{figure}[ht]
\centering
\includegraphics[scale=0.55]{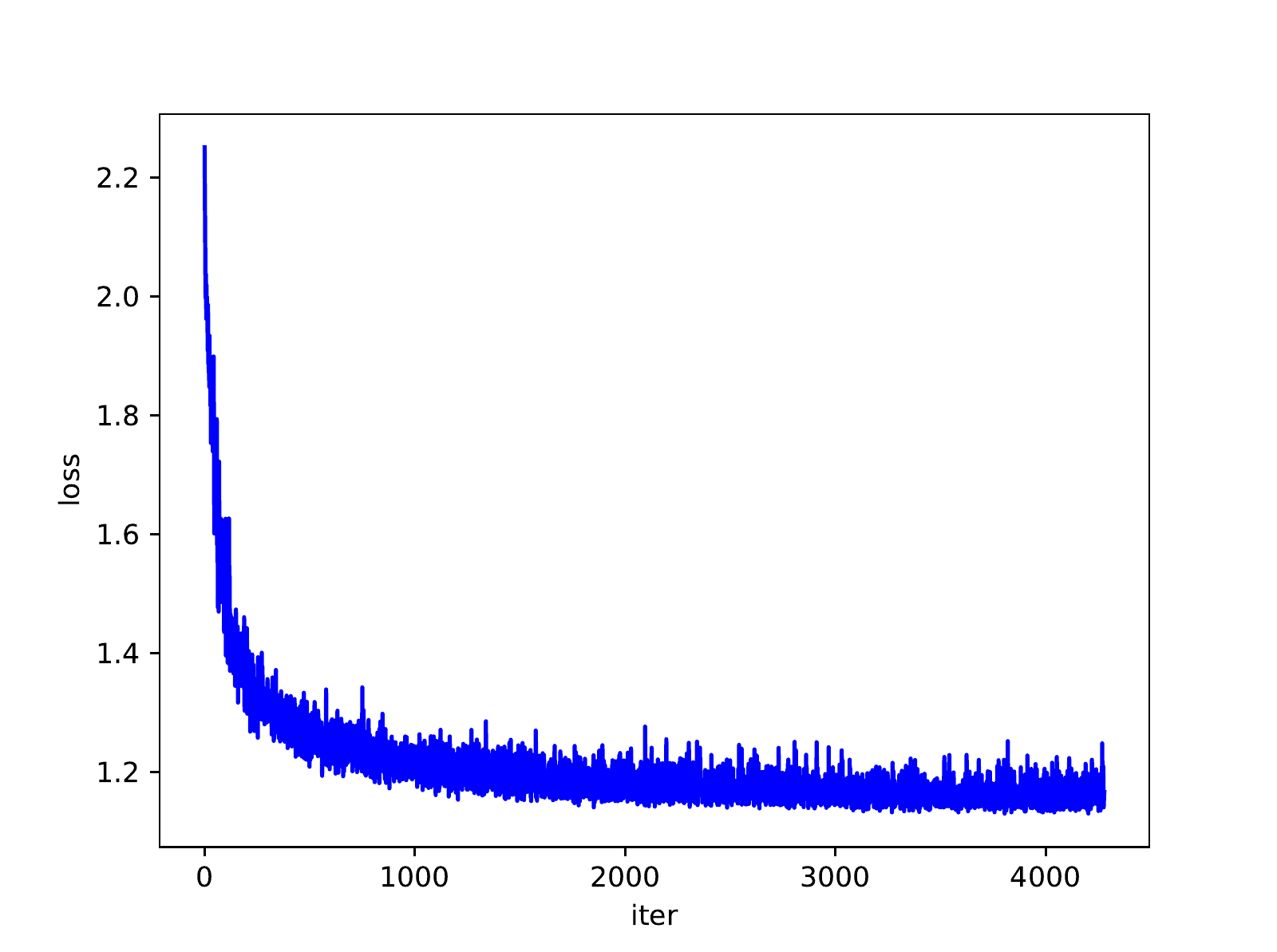}
\caption{Detailed training processes of the Dilation-Erosion strategy.}
\label{fig:lossCurve}
\end{figure}

\subsection{Convergence Analysis}

As a matter of fact, the dilation step may introduce noise labeling as any methods based on pseudo labels. To avoid error propagation, we locate a refined segment based on the STG algorithm to eliminate the background while keeping the action parts. To verify it, we conduct the convergence experiment and the result is shown in Fig. \ref{fig:lossCurve}. We can see that as the training iteration increases, the loss shows a gradual convergence trend. It proves that our method is able to avoid the error propagation problem.  

\begin{figure}[ht]
\centering
\includegraphics[scale=0.4]{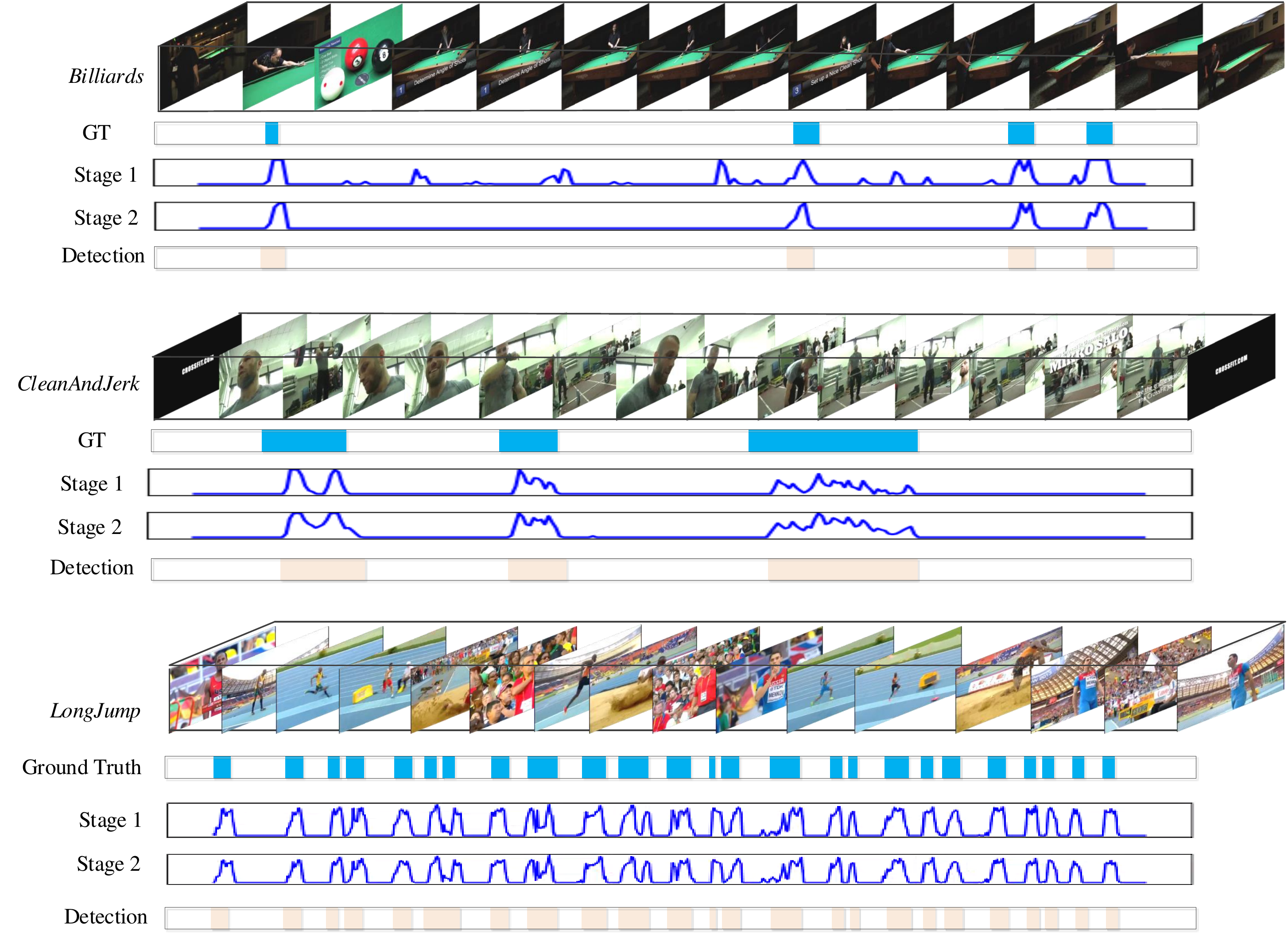}
\caption{Qualitative temporal action localization results on THUMOS14 dataset. For each example, we show the sampling frames, ground truth (GT), predictions of the stage 1 and stage 2 (the score is truncated at the threshold) and results of detection respectively. 
{\bf Top:} There are many background frames that are similar to action instances. Our method can identify the hard background, thereby suppressing background false positives. 
{\bf Middle:} The inter-frame similarity of the action instance is very low. With the help of dilation step, our method can alleviate incomplete movements.
{\bf Bottom:} There are dense action instances. Draw support from single frame annotation, dense action instances can be well distinguished.
The detection results prove that our approach has worked well in different situations.}
\label{fig:ANA1}
\end{figure}

\subsection{Qualitative Results}
In Fig. \ref{fig:ANA1}, three representative examples of qualitative TAL results illustrate the effectiveness of the proposed method. In the first example of the \emph{billiards}, in addition to shooting the billiards, the actor holds the billiard cue near the billiard table for most of the time, leading to many background false positives. The CAM of Stage One shows that many background frames are treated as action frames. However, with the help of Dilation-Erosion, it can be seen that the retrained model can suppress the generation of false positives. In the second example of the \emph{clean and jerk}, because there is an intra-class variety of frames in the action instance, the detections are incomplete as they are over-segmented in Stage One. After Stage Two, the less discriminative action parts are detected as well. In the third example of the \emph{long jump}, it shows that our method performs well when there are very dense action instances. These examples show that our method can effectively solve the problem of temporal action localization in complex environments.

\section{Conclusion}\label{sec5}
\label{conclusion}
In this paper, we explore using single-frame annotation for the task of temporal action localization. 
We propose a novel framework, which consists of the Snippet Classification model and the Dilation-Erosion module. 
The snippet Classification model can generate CAM and actionness score, which are used for inferring action segments  in a video and the following process of the Dilation-Erosion module. 
The Dilation-Erosion module is able to generate pseudo snippet-level labels, hard backgrounds and evident backgrounds, which in turn supervise the training of the Snippet Classification model. 
The two parts form a cyclic dependency.  During the two-stage learning process, the model can gradually improve the action completeness and effectively suppress false positives. 
We also propose a novel embedding loss that aggregates the features of action instances with the same label and separates the features of actions from backgrounds. 
Experiments show that the proposed framework even achieves comparable results compared with some fully-supervised methods on two standard benchmarks. 
Through mining the evident background as well as the hard background, the proposed method can obtain a better performance than SF-Net~\cite{10.1007/978-3-030-58548-8_25}.

\section{Declarations}
\textbf{Funding.} This work was supported by National Key R\&D Program of China (No. 2018AAA0102002), 
the National Natural Science Foundation of China (Grant No. 61672285, 62072245, and 61932020).

\noindent\textbf{Data Availability.} The datasets generated during and/or analysed during the current study are available in the GitHub repository: \url{https://github.com/LingJun123/single-frame-TAL}.


\bibliography{sn-bibliography}


\end{document}